\documentclass[11pt, a4paper, logo, copyright]{googledeepmind}

\usepackage[authoryear, sort&compress, round]{natbib}
\usepackage[utf8]{inputenc} % allow utf-8 input
\usepackage[T1]{fontenc}    % use 8-bit T1 fonts
\usepackage{hyperref}       % hyperlinks
\usepackage{url}            % simple URL typesetting
\usepackage{booktabs}       % professional-quality tables
\usepackage{amsfonts}       % blackboard math symbols
\usepackage{nicefrac}       % compact symbols for 1/2, etc.
\usepackage{microtype}      % microtypography
\usepackage{xcolor}         % colors
\usepackage{tcolorbox}
\usepackage{graphicx}
\usepackage{amsmath}
\usepackage{amssymb}
\usepackage{mathtools}
\usepackage{amsthm}
\usepackage{cleveref}
\usepackage{subcaption}

\DeclareMathAlphabet\mathbfcal{OMS}{cmsy}{b}{n}

\newcommand{\stateSpace}{\mathcal{X}}
\newcommand{\actionSpace}{\mathcal{A}}
\newcommand{\transFn}{\mathcal{P}}
\newcommand{\rewFn}{\mathcal{R}}
\newcommand{\reals}{\mathbb{R}}

\newcommand{\mdpTrain}{\mathbfcal{M}_{\textrm{train}}}

\newcommand{\mdpDeploy}{\mathbfcal{M}_{\textrm{eval}}}

\title{The Formalism-Implementation Gap in Reinforcement Learning Research}

\author[1]{Pablo Samuel Castro}

\affil[1]{Google DeepMind, Université de Montréal, Mila - Québec AI Institute}

\correspondingauthor{psc@google.com}

\begin{abstract}
The last decade has seen an upswing in interest and adoption of reinforcement learning (RL) techniques, in large part due to its demonstrated capabilities at performing certain tasks at ``super-human levels''. This has incentivized the community to prioritize research that demonstrates RL agent {\em performance}, often at the expense of research aimed at {\em understanding} their learning dynamics. Performance-focused research runs the risk of overfitting on academic benchmarks -- thereby rendering them less useful -- which can make it difficult to transfer proposed techniques to novel problems. Further, it implicitly diminishes work that does not push the performance-frontier, but aims at improving our understanding of these techniques. This paper argues two points: (i) RL research should stop focusing solely on demonstrating agent capabilities, and focus more on advancing the science and understanding of reinforcement learning; and (ii) we need to be more precise on how our benchmarks map to the underlying mathematical formalisms. We use the popular Arcade Learning Environment \citep[ALE;][]{bellemare2013ale} as an example of a benchmark that, despite being increasingly considered ``saturated'', can be effectively used for developing this understanding, and facilitating the deployment of RL techniques in impactful real-world problems.
\end{abstract}

\begin{document}

\maketitle

\section{Introduction}
\label{sec:intro}
Reinforcement learning (RL) has been a subject of research for at least 40 years, but it is only in the last decade where we have experienced a dramatic increase in its adoption beyond academic pursuits. The pivot point can likely be attributed to \citet{mnih2015humanlevel}, who introduced DQN (standing for {\bf D}eep {\bf Q}-{\bf N}etwork) which combined RL with deep neural networks and could play arcade games at super-human levels. The authors utilized the Arcade Learning Environment (ALE), which was introduced as ``a platform and methodology for evaluating the development of general, domain-independent AI technology'', presenting ``significant research challenges for RL'' \citep{bellemare2013ale}.

For a number of years the ALE served as a useful benchmark for evaluating how far we could push the performance of RL agents when trained for 200 million screen frames (around 38 hours of game play). Later, \citet{kaiser2020model} introduced the 100k benchmark (just under 2 hours of game play) to test the sample efficiency of RL agents. Even in the low-data regime of the 100k benchmark, there are already RL agents capable of achieving super-human performance \citep{schwarzer2023bbf}.
The fact that performance gains in the ALE are increasingly difficult to obtain have led to a general sense that the ALE has become a saturated and uninteresting benchmark.
As such, there is an implicit incentive for researchers to focus on other evaluation benchmarks where gains are less difficult to demonstrate. Unfortunately most of these benchmarks, including the ALE, lack a precise mapping to the underlying mathematical framework describing RL problems.

When viewed simply as a platform for evaluating agent {\em performance}, it is hard to argue against this somewhat defeatist perspective. However, it is worth recalling that the objective of most RL research is {\em not} to develop super-human video game players; rather, our hope is that the techniques and methods we develop may be useful for addressing impactful and challenging real-world problems. To do so effectively would require a robust understanding of how to transfer techniques developed on one set of environments to new, and likely very different, ones. Following the {\em insight-oriented exploratory research} advocated by \citet{herrmann2024rethink} and the {\em scientific testing} suggested by \citet{jordan2024benchmarking}, RL research should focus on developing this understanding.

Transferring an algorithm evaluated on one benchmark to another environment can be rather complex. As \citet[Ch. 16]{sutton98rl} writes: ``Applications of reinforcement learning
are still far from routine and typically require as much art as science. Making applications
easier and more straightforward is one of the goals of current research in reinforcement
learning''; a statement that is still true almost thirty years later. Yet, there are a growing number of examples of RL being used successfully in real-world applications (see \cref{app:realWorldRL} for a list of some examples); however, most (if not all) of these have required employing a team of RL experts to achieve success.

\begin{figure*}[!t]
    \centering
    %\raisebox{0.2\height}
    {\includegraphics[width=0.8\textwidth]{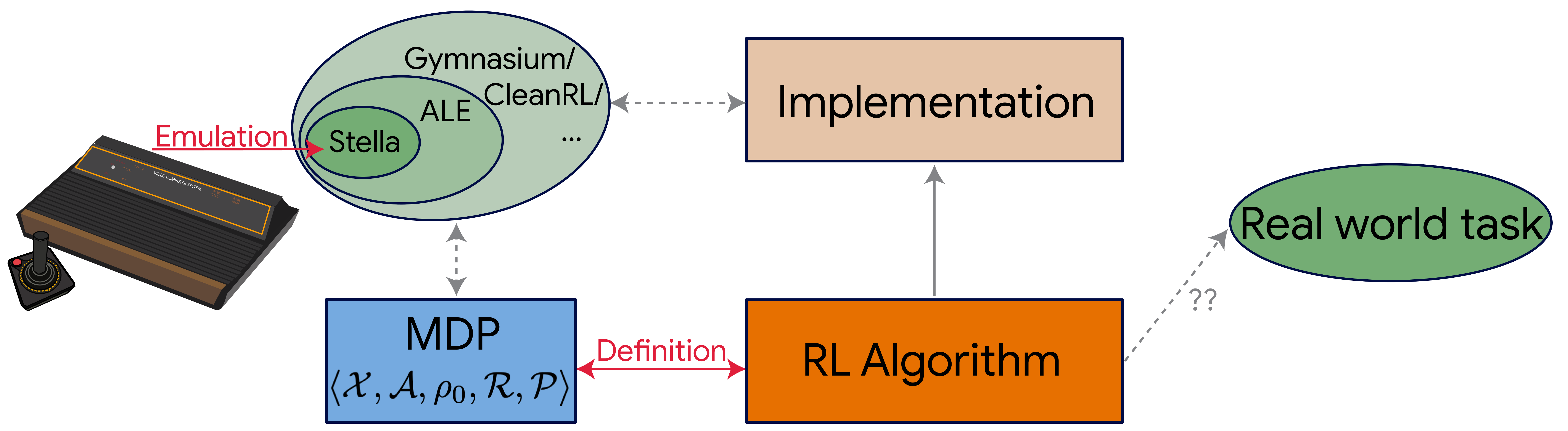}}
    %\vspace{-0.1cm}
    \caption{RL Algorithms are typically defined using the mathematical formalism of MDPs, but implemented (somewhat) independently with code, and evaluated on standard benchmarks such as the ALE. If we want to maximize the transferability of our algorithms, {\bf we need to be explicit about the MDP-benchmark and algorithm-implementation mappings}.}
    %\vspace{-0.2cm}
    \label{fig:mdpMapping}
\end{figure*}

This lack of transfer may be due to the fact that as a community we have over-indexed on environment-centric concepts, rather than agents (the ``environment spotlight dogma'' \citep{abel2024dogmas}). A complementary perspective is that we have focused on demonstrating agent capabilities (i.e. how well it performs against a baseline) as opposed to {\em how} and {\em why} an agent performs the way it does. Successful transfer would require properly understanding both (1) how algorithmic specifications (defined in abstract, mathematical terms) transfer to environment implementations (which include a number of non-trivial implementation design choices), and (2) the sensitivity of an agent to varying hyper-parameters and environmental properties (see \cref{fig:mdpMapping}).

{\bf This paper argues that, in order to develop transferable RL techniques, we need to be more formally explicit with benchmark specifications, and we need to stop focusing solely on performance gains.} This shift in focus can help facilitate the use of reinforcement learning in more challenging and impactful real-world problems. 

\section{The formalism-implementation gap}
\label{sec:closingGap}

RL algorithms are typically presented abstractly using the mathematical formalism of Markov decision processes, which enables their potential application to a wide variety of problems. These algorithms are then compared to baselines on a standard benchmark such as the ALE with the implication that stronger performance would transfer to other problems of interest. As \cref{fig:mdpMapping} illustrates, this requires explicit MDP-benchmark and algorithm-implementation mappings. Although the latter mapping is discussed more frequently, implementations necessarily involve a set of hyper-parameters to tune, which are not always broadly explored. As a number of works have demonstrated, an algorithm's performance is often mostly due to hyper-parameter settings rather than its mathematical properties. For instance, \citet{henderson2018deep} demonstrated how sensitive algorithms can be to choices of network architectures, reward scaling, and implementations; while \citet{kaiser2020model} claimed their method (SimPLe) outperformed Rainbow \citep{hessel2018rainbow} in the Atari 100k benchmark, \citet{vanHasselt2019when} demonstrated that with proper tuning, Rainbow was significantly stronger; \citet{obando2021revisiting} showed significant improvements in DQN by simply switching the optimizer from RMSProp (as used by \citet{mnih2015humanlevel}) to Adam \citep{kingma2015adam}; and \citet{ceron2024on} performed an exhaustive evaluation of various hyper-parameters, convincingly demonstrating how much they affect performance and algorithmic rankings.

\subsection{Markov decision processes}
\label{sec:mdps}
A Markov decision process (MDP) is a tuple $\mathcal{M}:=\langle \stateSpace, \actionSpace, \rho_0, \rewFn, \transFn \rangle$, where $\stateSpace$ and $\actionSpace$ denote the state and action spaces, respectively; $\rho_0\in\Delta(\stateSpace)$ (for any set $C$, $\Delta(C)$ denotes a distribution over $C$) is a distribution over start states; $\rewFn:\stateSpace\times\actionSpace\rightarrow\reals$ is a reward (or cost) function; and $\transFn:\stateSpace\times\actionSpace\rightarrow\Delta(\stateSpace)$ is the transition function, where $\transFn(x, a)(x')$ denotes the probability of transitioning from state $x$ to state $x'$ under action $a$.

An RL agent is traditionally considered a separate {\em entity} that interacts with the MDP and produces a behaviour policy $\pi:\stateSpace\rightarrow\Delta(\actionSpace)$. The {\em value} of a policy $\pi$ is defined as the expected sum of discounted rewards:
\[ V_{\gamma}^{\pi} = \mathbb{E}_{\pi}\left[ \sum_{t=0}^{\infty} \gamma^t \rewFn(x_t, a_t) \right] \]
where $s_0\sim\rho_0, a_t\sim\pi(s_t)$, $s_{t+1}\sim\transFn(s_t, a_t)$, and $\gamma\in [0, 1)$ is a discount factor. Note that many works include $\gamma$ as part of the MDP definition. The aim of RL agents is to find the optimal policy, defined as $\pi_{\gamma}^* = \arg\max_{\pi}V_{\gamma}^{\pi}$, which is guaranteed to exist \citep{puterman2014markov}. Note that the value of a policy (and by extension its optimality) depends on $\gamma$, which suggests that optimality {\em is an agent-centric notion}, relative to $\gamma$. Indeed, if the same algorithm is run with two different discount factors $\gamma_1$ and $\gamma_2$, it is not clear which of $\pi^*_{\gamma_1}$ or $\pi^*_{\gamma_2}$ is ``better''. On the ALE, this is somewhat addressed given that performance comparisons are done by reporting average cumulative {\em undiscounted} rewards for a finite horizon; we discuss this further in \cref{sec:empiricalEvaluations}.

A more general notion is that of Partially-Observable MDPs (POMDPs) which also consider a finite set of observations $\Omega$ and an observation function $\mathcal{O}:\stateSpace\times\actionSpace\rightarrow\Delta(\Omega)$, where $\mathcal{O}(x, a)(\omega)$ denotes the probability of observing $\omega\in\Omega$ when taking action $a$ in state $x$ \citep{kaelbling1998planning}. In this setting, an agent does not have access to the true underlying state, but must infer it from the observations it received. Importantly, whereas in an MDP the agent can act optimally in a Markov fashion (i.e. optimal policies need only be functions of the current state), in POMDPs this is not generally the case: policies that are functions purely of the current observations are not optimal \citep{singh1994learning}. There are a variety of approaches for dealing with partial-observability, such as maintaining a {\em belief} over the states: $b\in\Delta(\stateSpace)$, or having policies be functions of trajectories of finite length: $\pi:\stateSpace^H\rightarrow\Delta(\actionSpace)$.

To formalize the preamble of this section, an RL algorithm/agent is a process $A$ that receives a (PO)MDP $\mathcal{M}$ with which it can interact and produces a policy $\pi$ that maximizes cumulative rewards. This includes learning both online (where the agent collects data from $\mathcal{M}$ by directly interacting with it) and offline (where the agent only has access to a pre-generated dataset of interactions with $\mathcal{M}$). An RL agent that has been demonstrated to produce good performance on a set of (PO)MDPs $\mdpTrain$ should ideally be able to also produce good performance in a new MDP $\mdpDeploy$.

\begin{tcolorbox}[colback=blue!10, colframe=blue!65!black, title=\textbf{Recommendation 1}]
Be explicit about the mapping between formalism and implementation.
\end{tcolorbox}

\section{The ALE as a POMDP}
\label{sec:alePOMDP}

In this section we use the ALE to illustrate the importance of \textcolor{blue}{Recommendation 1} and how loose this mapping is in practice. Our intent with this section is not to be prescriptive, but rather to highlight the cost of {\em not} being explicit with these mappings when drawing conclusions about algorithms.

The ALE serves as an interface to Stella\footnote{http://stella.sourceforge.net/}, which is an emulator of the original Atari 2600 gaming system, thereby providing access to hundreds of Atari games. In the original Atari 2600 system users played most games with a joystick, which triggered one of 9 position events, as well as a ``fire'' event, for a total of 18 possible events\footnote{Other controller types were available, such as paddles, but were less common \citep{montfort09racing}.}. RL agents thus interact with these Atari games by sending actions to the ALE, which transmits these to Stella to update the game state; after executing the selected actions, the ALE will extract the relevant information from the updated game state (i.e. screen setting, reward received, end-of-life signals, etc.) and transmit these back to the RL agent. %This seems to follow the traditional agent-environment dichotomy presented in most introductory textbooks \psc{see Figure X, left}, which should in turn result in a natural mapping from an Atari game to an MDP.

A natural, but na{\" i}ve, mapping considers the set of possible screen states as $\stateSpace$, the set of possible joystick (or paddle) positions as $\actionSpace$, and the Stella emulator itself (which updates screen states) as $\transFn$. In practice, however, the mapping is not this simple.
%; what is worse, the separation between agent and environment is not even clear. 
The first reason is that the ALE itself provides a series of options that affect how it handles communication between the RL agent and Stella. The second reason is that most RL agents do not interact directly with the ALE; rather, researchers use existing software libraries (such as Gymnasium \citep{towers2024gymnasium}, StableBaselines3 \citep{stable-baselines3}, CleanRL \citep{huang2022cleanrl}, or Dopamine \citep{castro2018dopamine}), as they provide a common interface to a variety of other popular benchmarks. These libraries provide their own set of options which further affects the communication between the RL agent and Stella, and complicates the agent-environment divide
(\cref{fig:mdpMapping}). These options form a part of our MDP-mapping discussion below, and we provide a more comprehensive list in \cref{app:stellaWrapperOptions}.

\begin{figure*}[!t]
    \centering
    %\raisebox{0.2\height}
    {\includegraphics[width=0.97\textwidth]{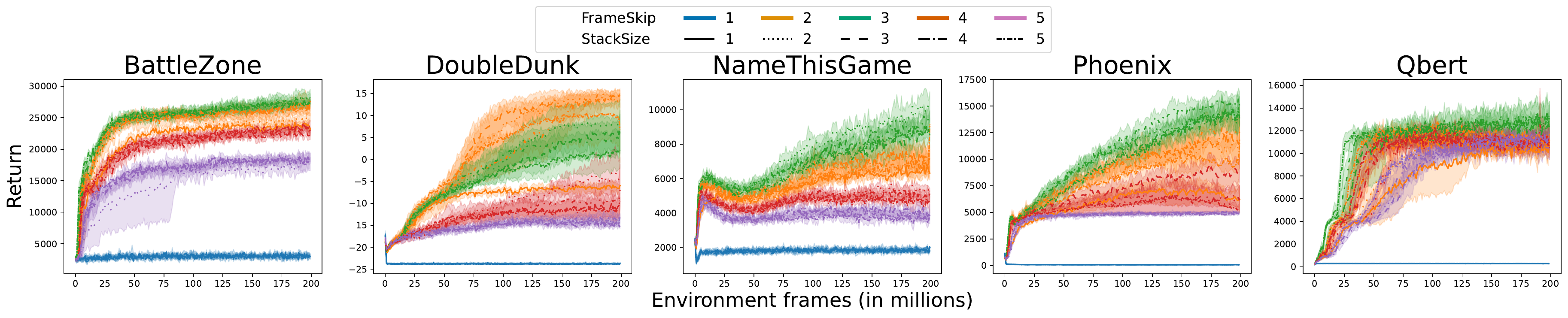}}
    %\vspace{-0.1cm}
    \caption{The impact of {\bf frame skip} and {\bf frame stacking} on DQN. Evaluated on Dopamine \citep{castro2018dopamine} on the Atari-5 games \citep{aitchison2023atari5}; solid lines are mean performance over 5 independent seeds, shaded areas represent 95\% confidence intervals.}
    %\vspace{-0.2cm}
    \label{fig:fs_ss_sweep}
\end{figure*}

\subsection{The state space $\stateSpace$}
\label{sec:aleStateSpace}

There are a few options that can affect the meaning of ``state'', but two of the most prominent ones are frame skipping and frame stacking. With a {\bf frame skip} value of $n$, the action sent to Stella will be {\em repeated} $n$ times\footnote{Note that \citet{bellemare2012investigating} and \citet{bellemare2013ale}  originally used $n=5$, but \citet{mnih2015humanlevel} changed this to $4$, which is now the most commonly used setting.}. Typically, there is also a max pool operation applied on the last two frames.  If the observation returned from the ALE is a tensor of shape $[w, h, d]$, {\bf frame stacking} takes the last $m$ frames (as received by the agent) and stacks them to create a new tensor $[m, w, h, d]$ which is fed to the agent as input. As \cref{fig:fs_ss_sweep} shows, these choices can have a big impact on performance.

The current default values for both frame stacking and skipping are $4$, with a max pool over the last 2 skipped frames. Thus, when an RL agent takes an action at environment step $t$, what is returned from the environment is not frame $x_{t_1}$, but rather $\textrm{maxpool}(x_{t+3},x_{t+4})$ (due to frame skipping), and the agent state that is fed to the neural network would be something akin to:
\[ 
\left[ \textrm{maxpool}(x_{t-10},x_{t-9}),  
  \textrm{maxpool}(x_{t-6},x_{t-5}),
  \textrm{maxpool}(x_{t-2},x_{t-1}),
  \textrm{maxpool}(x_{t+3},x_{t+4}) 
\right]
\]

This implies that most RL agents {\em are not acting in a purely Markovian fashion}; the fact that changing the number of frames stacked and/or skipped has such a large impact on performance also suggests that {\em the ALE is not an MDP}, at least not in the na{\" i}ve mapping being considered.

A more accurate statement would thus be that each of the frames returned by the ALE are in fact {\em observations} $o_t$ and the ALE should really be considered as a POMDP, with each of these processing steps as techniques to deal with partial observability. \citet[Appdx]{mnih2015humanlevel} do partially address this by considering the full sequence of aliased state-action observations as the state: $x_t = o_1,a_1,o_t,a_2,\ldots,a_{t-1},o_t$, which leads to a fully Markovian system; however, the use of the full sequence is purely a theoretical construct and not what is done in practice.

\begin{figure}[!t]
    \centering
    \begin{subfigure}[t]{0.45\textwidth}
        \centering
        {\includegraphics[width=0.6\textwidth]{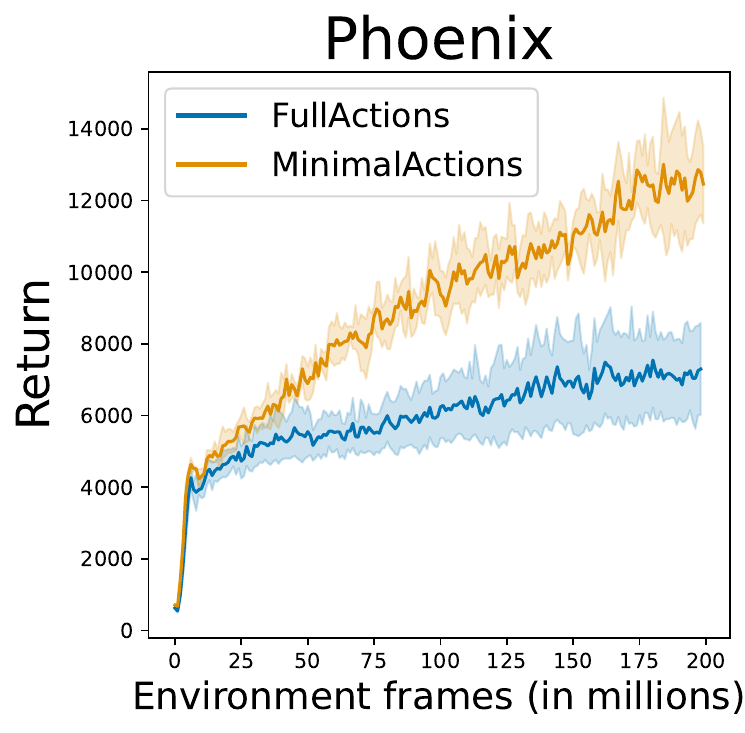}}
        \caption{Minimal action sets.}
        \label{fig:minimalActions}
    \end{subfigure}
    ~
    \begin{subfigure}[t]{0.45\textwidth}
        \centering
        {\includegraphics[width=0.6\textwidth]{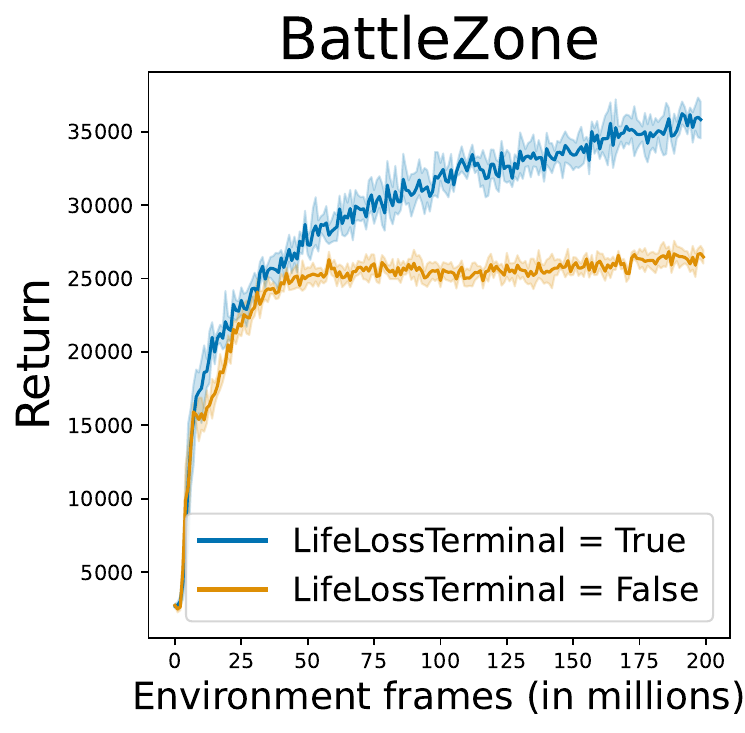}}
        \caption{Life-loss events as episode termination.}
        \label{fig:lifeLoss}
    \end{subfigure}
    \caption{Varying the definition of $\actionSpace$ and $\transFn$ can have a large impact on reported performance. Evaluated on Dopamine \citep{castro2018dopamine} with DQN, solid lines are mean performance over 5 independent seeds, shaded areas represent 95\% confidence intervals.}
\end{figure}

\subsection{The action space $\actionSpace$}
\label{sec:aleActionSpace}
An important option made available by the ALE is the option to use a minimal action set, which is turned on by default in most libraries. In some games certain joystick positions are effectively no-ops; for instance, in \texttt{Phoenix} the only valid moves are \texttt{Left,Right,Down} and \texttt{Fire}, which means that the \texttt{DownLeft, Left}, and \texttt{UpLeft} events all map to the same \texttt{Left} event. In these games, the ALE reduces the action space from the full 18 set to a smaller number (e.g. $8$ for \texttt{Phoenix}).

Two popular games in the ALE -- Pong and Breakout -- were originally played with paddles as opposed to joysticks. While joysticks only trigger discrete events, paddles were fully continuous. In the original implementation by \citeauthor{bellemare2013ale}, the movement of the paddles were mapped onto discrete \texttt{Left,Right} actions, effectively projecting the game paddles onto a smaller set of 3 possible positions\footnote{See \url{src/ale/environment/ale_state.hpp.}}. Although \citet{farebrother2024cale} recently added support for continuous actions, they did not yet include support for paddles.

Two implication of these options are that (1) they can result in a set $\mdpTrain$ with inconsistent action sets and (2) they can change the difficulty of the games from what they were originally meant to be, as illustrated in \cref{fig:minimalActions}.

\subsection{The initial state distribution $\rho_0$}
\label{sec:aleInitialStateDistribution}
The Atari 2600 system, and Stella by extension, are deterministic systems by design. To avoid overfitting, \citet{mnih2015humanlevel} added a series of no-op actions at the start of each episode, where the number of no-ops is selected uniformly randomly between 0 and \texttt{max\_noop} (\citet{mnih2015humanlevel} used \texttt{max\_noop = 30}, which has remained the default in most implementations). This results in a non-trivial, and more interesting, start-state distribution, but does step away from the original design.

\subsection{The reward function $\rewFn$}
\label{sec:aleRewardFunction}
Given that the reward magnitude varies widely across Atari 2600 games, \citet{mnih2015humanlevel} clipped all rewards at $[-1, 1]$, which enables the use of global hyper-parameters across games (such as learning rate), and facilitates numerical optimization. This design choice can have an aliasing effect (e.g. the agent sees no difference between a reward of $1$ and a reward of $1000$, even though they may result from very different game states), resulting in partially observable rewards (in addition to states).

\subsection{The environment dynamics $\transFn$}
\label{sec:aleTransFunction}
Transitions are defined between the ground states of an MDP, so the definition of $\transFn$ depends on what $\stateSpace$ is taken to be. As discussed in \cref{sec:aleStateSpace} it is more accurate to consider the ALE as a partially observable system, where the agent does not have access to the true states. This may have implications for methods that learn world models, in particular if they include theoretical considerations.

Further, there are design decisions made by the ALE wrappers (such as Dopamine, Gymnasium, and CleanRL) which also affect the transition dynamics as experienced by the agent. For instance, games are reset upon episode termination, which typically means a $\textrm{GameOver}$ signal; when games are reset, the agent state is reinitialized according to $\rho_0$ (see \cref{sec:aleInitialStateDistribution}). Certain games provide the player with multiple lives, and one can set an option to terminate episodes when life-loss events occur, which directly impacts the environment dynamics, as perceived by the agent. Figure~\ref{fig:lifeLoss} demonstrates that this option can have a large impact on downstream agent performance. Frustratingly, this option defaults to \texttt{False} in Dopamine \citep{castro2018dopamine} and Gymnasium \citep{towers2024gymnasium}, but defaults to \texttt{True} in CleanRL \citep{huang2022cleanrl} and StableBaselines3 \citep{stable-baselines3}.

\begin{figure*}[!t]
    \centering
    %\raisebox{0.2\height}
    {\includegraphics[width=\textwidth]{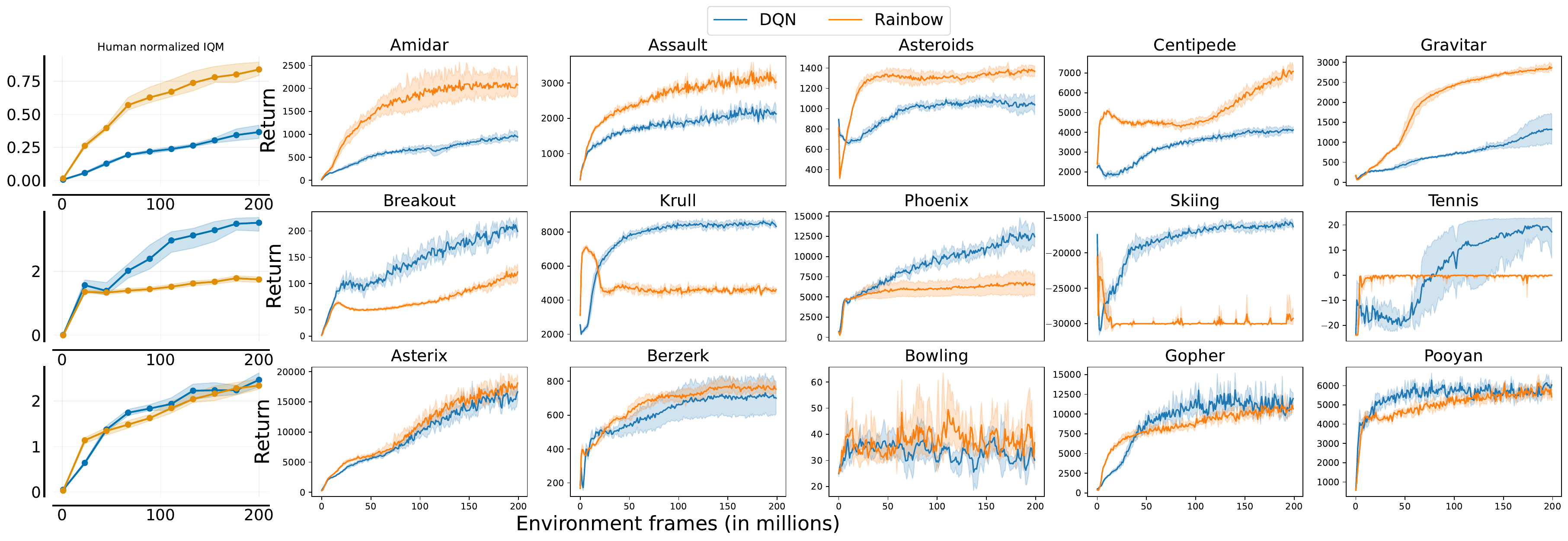}}
    %\vspace{-0.1cm}
    \caption{Evaluating the comparison of DQN and Rainbow \citep{hessel2018rainbow} over different subsets of games. In each row, the leftmost column depicts the human-normalized Interquartile mean with 95\% stratified bootstrap confidence intervals \citep{agarwal2021deep}, computed over the games on the same row, with 5 independent seeds for each. {\bf The ranking of the algorithms can dramatically change depending on the subset of games chosen.}}
    %\vspace{-0.2cm}
    \label{fig:gameSubsets}
\end{figure*}

\section{Measuring progress}
\label{sec:empiricalEvaluations}

Given the increasingly empirical nature of reinforcement learning research, progress is typically quantified via accumulated rewards. A number of decisions in the evaluation procedures can dramatically affect the conclusions drawn from the presented results, and thereby their generality to novel scenarios. We once again utilize the ALE as an illustrative example for the points made below.

\paragraph{Choice of discount factor}

Although $\gamma$ plays a central role in most RL algorithms, it is not included in the reported performance metrics. Instead what is reported is the average cumulative {\em undiscounted} rewards per episode throughout training, where an episode lasts until \texttt{max\_steps\_per\_episode} agent steps have elapsed, or until an \texttt{end\_of\_episode} signal is received. While effective at enabling comparisons across various algorithms (for instance those using different values of $\gamma$), it is {\em a different objective than what most RL agents are optimizing for}. This is perhaps desirable (it can act as a form of test setting), but it surely deserves more consideration than it currently receives. One solution to this ambiguity is for research papers to explicitly report $\{ \gamma_{\textrm{train}} \}$ and $\gamma_{\text{eval}}$ (e.g. for the ALE we would normally set $\gamma_{\text{eval}} = 1$); note that we indicate $\{ \gamma_{\textrm{train}} \}$ as a set, as some works propose training agents with multiple discount factors \citep{fedus2019hyperbolicdiscountinglearningmultiple,romoff2019separating}. It has also been argued that discounting is ``fundamentally incompatible with function approximation for control in continuing tasks'' \citep{naik2019discounted}.

\begin{tcolorbox}[colback=blue!10, colframe=blue!65!black, title=\textbf{Recommendation 2}]
Be explicit about $\{ \gamma_{\textrm{train}} \}$ and $\gamma_{\text{eval}}$ when presenting  results. While $\{ \gamma_{\textrm{train}} \}$ may vary, $\gamma_{\text{eval}}$ must always be consistent between compared algorithms.
\end{tcolorbox}

\paragraph{Choice of experiment length}
The standard experiment length proposed by DQN was 200 million environment frames (or 50 million agent steps, due to frame skipping). Since then researchers have used a variety of different experiment lengths, ranging from the 100k benchmark (100 thousand agent steps or 400 thousand environment frames) proposed by \citet{kaiser2020model}, to 4 million \citep{farebrother2023protovalue}, 40 million \citep{graesser2022state}, and 100 million \citep{ceron2023small} environment frames. While some of these choices may be driven by the research in question, they are most likely driven by computational considerations (as well as some arbitrariness). \citet{ceron2024on} demonstrated that the ranking of algorithms can be highly sensitive to the choice of experiment length. As argued by \citet{patterson2024empirical}, the question of experiment length should be driven by the scientific question being asked. For instance, even though focused on performance on the 100k benchmark, \citet{schwarzer2023bbf} ran their BBF method for longer to ensure it was not overfit to the training regime and could continue to improve. Finally, the choice of experiment length may also be driven by the choice of $\{ \gamma_{train}\}$ and $\gamma_{eval}$, as discussed above.

\begin{tcolorbox}[colback=blue!10, colframe=blue!65!black, title=\textbf{Recommendation 3}]
Be explicit about experiment length, and how it connects to the research in question.
\end{tcolorbox}

\paragraph{Choice of $\mdpTrain$ and $\mdpDeploy$}
In addition to the evaluation proposals put forth by \citet{bellemare2013ale}, there have been a number of subsequent methodologies suggested by followup works \citep{machado2018revisiting,agarwal2021deep,patterson2024empirical}, as well as paper-specific evaluations, which can sometimes lead to inconsistencies in results. In general, however, the community has largely converged towards reporting human-normalized scores\footnote{``Human'' refers to the test gamer employed by \citet{mnih2015humanlevel}.}, aggregated over multiple games and seeds. The ranking of algorithms is then a simple comparison of the final aggregate performance (see left column of \cref{fig:gameSubsets}).

Ideally, a subset of games ($\mdpTrain$) is used to tune algorithms  (e.g. the five game {\em training set} of \citet{bellemare2012investigating}) and a separate set of games ($\mdpDeploy$) are used to evaluate them. This practice has become less common, with results presented on a single set of games (i.e. $\mdpTrain \equiv \mdpDeploy$).
Further, only a subset of the full game set is often used to reduce the computational burden; unfortunately, the subsets chosen are quite varied, which can dramatically affect rankings. Indeed, in Figure~\ref{fig:gameSubsets} we present aggregate results over different choices of $\mdpTrain$ and observe dramatically different aggregate results. To try and remedy this, \citet{aitchison2023atari5} proposed 5- and 10-game subsets, based on a statistical approach that strives for order-preservation, which are predictive of final performance. These subsets are still focused on performance, however, and use a summary measure obtained via aggregation.

\begin{tcolorbox}[colback=blue!10, colframe=blue!65!black, title=\textbf{Recommendation 4}]
Be explicit about $\mdpTrain$ and $\mdpDeploy$, which are ideally disjoint. Only compare against other algorithms with $\mdpDeploy$.
\end{tcolorbox}

\paragraph{Avoid aggregate results}
A number of recent works have highlighted how sensitive most algorithms are to the choice of hyper-parameters \citep{vanHasselt2019when,ceron2024on,adkins2024method,patterson2024empirical}, which further calls into question any proposed ranking, let alone the generality of algorithmic findings. Indeed, \citet{ceron2024on} convincingly demonstrated that RL agents are quite sensitive to hyper-parameter selection, and this sensitivity (and induced ranking) varies widely from environment to environment. The same authors also demonstrated that algorithmic rankings can be sensitive to the number of episodes run. As argued by \citet{patterson2024empirical}, the choice of how long to train should depend on the scientific question being asked, and whether the learning curves suggest the algorithm has converged; this choice will likely vary between games. Indeed, any evaluation focusing on aggregate performance is failing to see the trees for the forest, to flip the common adage.

\begin{tcolorbox}[colback=blue!10, colframe=blue!65!black, title=\textbf{Recommendation 5}]
Reduce focus on aggregate agent performance and opt for per-game analyses.
\end{tcolorbox}

\section{Desiderata for benchmarks}
Our goal as RL researchers should shift from designing SotA-claiming agents to developing methods and insights that are well-specified, consistent across different training regimes, robust to varying hyper-parameters, and without sacrificing performance. Even under this perspective, benchmarks have an important role to play in aiding progress.
In this section we specify some desiderata for what benchmarks should strive for and argue that the ALE, despite being increasingly considered a ``saturated'' benchmark, remains useful for scientific progress.

\subsection{Benchmarks should be well understood}
Given that most research is evaluated by a diverse set of reviewers, it is important that this diversity of readers has an adequate level of understanding of the challenges and nuances of the benchmark. Familiarity facilitates the evaluation and critique of new research, which is vital to scientific progress. This may be one of the reasons why the ALE remains one of the most used benchmarks. It is also worth remarking that researchers are more likely to be familiar with video games than with, for instance, stratospheric balloon control \citep{greaves2021ble}.

\subsection{Benchmarks should be diverse and avoid experimenter-bias}

A suite of diverse environments yields flexibility in selecting $\mdpTrain$ and $\mdpDeploy$. Ideally, this suite of environments was developed {\em independently} of RL objectives to avoid faulty reward functions or ``reward hacking'' \citep{clark2016faulty}. This form of experimenter bias is unfortunately present in many recent environment suites that have been developed specifically for RL research.

The ALE suite consists of around 60 games which vary in difficulty, reward sparsity, observational complexity, and game dynamics. Importantly, these games were developed by professional game designers for human enjoyment, and not by RL researchers for RL research. Many of these games also include difficulty modes which are useful for  a number of research topics, including generalization and curriculum learning \citep{farebrother2018generalization}.

As previously mentioned, Atari games were originally deterministic but \citet{machado2018revisiting} introduced ``sticky actions'', which causes actions to repeat (or ``stick'') with some (controllable) probability, thereby rendering the environment transitions stochastic. This can affect the difficulty of games and can avoid trivial open-loop policy solutions, in particular when coupled with \texttt{no-op} restarts (see \cref{sec:aleInitialStateDistribution}).

Since its introduction, the ALE has been a suite of discrete action environments, yet the original joystick was an analogue controller. \citet{farebrother2024cale} introduced the Continuous ALE (CALE), which enables continuous actions for the ALE. Rather than having 9 discrete positions for each joystick position, the actions are parameterized via three continuous dimensions. The joystick sensitivity can be controlled via a parameter, which does affect downstream agent performance. As stated by the authors, the fact that both the ALE and the CALE share the same underlying emulator enables a direct comparison between discrete- and continuous-action agents.

\subsection{Benchmarks should be naturally extendable}
It is unlikely that all benchmarks will capture all downstream uses when investigating new agent features and capabilities. However, introducing an entirely new benchmark for the purpose of evaluating a specific capability runs the risk of introducing experimenter bias. It is thus ideal if benchmarks were extensible so as to enable novel research without sacrificing familiarity.

The ALE serves as a useful example for these types of extensions, and we outline a few of them below. \citet{shao2022MaskAF} added support for masked observations, which can be useful for exploring ideas related to partial observability. This can be useful for clarifying the MDP-benchmark mapping discussed in \cref{sec:alePOMDP}.  \citet{delfosse2024ocatari} proposed OCAtari, an object-centric variant of the ALE which can enable investigating how agents learn with varying levels of semantic information. \citet{terry2020MultiplayerSF} added multi-player support to the ALE, which is useful for multi-agent RL research.
\citep{young2019minatar} proposed Miniature Atari (MinAtar) as an Atari-inspired benchmark that can be trained at a fraction of the compute. \citet{gymnax2022github} provides a Jax-optimized version of this suite, which can be used for high-throughput training. Similarly, \citet{dalton2020accelerating} introduced a GPU-Accelerated Atari emulation for faster training.
As mentioned above, \citet{aitchison2023atari5} proposed the Atari-5 and Atari-10 subsets as representative of the full suite. In a similar vein, \citet{obando2021revisiting} argued that, if carefully designed, experiments on smaller environments such as the set of ``classic control'' environments \citep{brockman2016openai} or MinAtar can yield insights which transfer to the full ALE suite. These approaches can be useful for reducing the computational burden of RL research.

\begin{tcolorbox}[colback=blue!10, colframe=blue!65!green, title=\textbf{Recommendation 6}]
Prioritize benchmarks which are well-understood, diverse, avoid experimenter-bias, and are naturally extensible.
\end{tcolorbox}

\section{Conclusions and concrete suggestions}

If our only goal is to ``play Atari games well'', then perhaps the ALE has indeed outlived its usefulness. However, that is not the purpose of academic benchmarks: they exist to advance science. As such, we should be striving to develop methods that are reliable and transferable, and this requires going beyond pure ``SotA-chasing'', aggregate cumulative returns, and simplistic algorithmic rankings; we should be more focused on fine-grained analyses to get a better sense of the underlying characteristics and pathologies of reinforcement learning methods. An important part of this approach is to be explicit about how the benchmarks used map to the underlying mathematical formalisms.

Throughout this paper we provided a number of recommendations for improving the science of reinforcement learning. We elaborate on a few concrete suggestions for following these guidelines.

One of the main recommendations of this paper is to be explicit about the formalism-implementation mapping (\textcolor{blue}{Recommendation 1}). A number of things can help in this regard, such as
providing open-source code, specifying which version of the libraries were used for the experiments, and providing a listing of all the hyper-parameters. Notably, be explicit about the (PO)MDP specification of the environments (\cref{sec:alePOMDP}), $\{ \gamma_{\textrm{train}} \}$ and $\gamma_{\text{eval}}$ (\textcolor{blue}{Recommendation 2}), experiment length (\textcolor{blue}{Recommendation 3}), and $\mdpTrain$ and $\mdpDeploy$ (\textcolor{blue}{Recommendation 4}).

When analyzing results, do so at the game level (\textcolor{blue}{Recommendation 5}), and focus on addressing hypotheses as opposed to claiming ``superiority''. What are the games where performance gains are observed? Are there patterns in these games? Do they correlate with per-game results from prior works? Are these per-game results sensitive to hyper-parameter choices? Recall that the point of reported metrics are to serve as a signal for the underlying characteristics of the method in question. As such, the more signals we can provide, the richer the understanding we will have of the underlying algorithms. This includes, but is not limited to, reporting confidence intervals, $p$-values, and full return distributions. See \citet{agarwal2021deep} and \citet{patterson2024empirical} for guidance on this topic.

Being explicit about environment and benchmark specifications can help avoid the ``environment spotlight dogma'' \citep{abel2024dogmas}. This specificity may also have consequences for the ``mirror of empowerment'' perspective, and can help address the ``tension between  the plasticity and empowerment of an agent'' \citep{abel2025plasticitymirrorempowerment}. 

It can be argued that in a world where AI models are being used by millions of people worldwide, mostly due to large language models (LLMs), video games in general have lost their relevance for research. To this we refute that the success of LLMs rests on decades of academic research that was not premised on its application to these use-cases. Indeed, we argue that over-indexing on a single flavor of research, whether it be LLMs or reinforcement learning from human feedback (RLHF), has an opportunity cost in terms of potential, and unanticipated, breakthroughs. Further, we argue that when benchmarks are approached for insight-oriented exploratory research \citep{herrmann2024rethink} and scientific testing \citep{jordan2024benchmarking}, the results are likely applicable and relevant to related fields such as RLHF and LLMs (\textcolor{blue}{Recommendation 6}).

In summary, this paper has argued that shifting our focus towards building scientific insights via a well-specified set of evaluation benchmarks will yield more transferable, and useful, RL algorithms.

\paragraph{Acknowledgements}
The author would like to thank David Abel, Joao Madeira Araujo, Tom Schaul, Doina Precup, Georg Ostrovski, Will Dabney, Johan Obando Ceron, Roger Creus-Castanyer, Dhruv Sreenivas, Olya Mastikhina, Julian Dierkes, and the many RL researchers with which I've discussed many of these topics over the years.

\bibliography{references}

%%%%%%%%%%%%%%%%%%%%%%%%%%%%%%%%%%%%%%%%%%%%%%%%%%%%%%%%%%%%

\appendix
\newpage

\section{Stella/ALE wrapper options}
\label{app:stellaWrapperOptions}

\begin{table}[!h]
\caption{Stella-controlled options, as provided by the ALE constructor. Taken from the object constructor in https://github.com/Farama-Foundation/Arcade-Learning-Environment/blob/master/src/ale/python/env.py.}
\label{tbl:stellaOptions}
\vskip 0.15in
\begin{center}
\begin{small}
%\begin{sc}
\begin{tabular}{p{3.5cm}p{1.5cm}p{3.5cm}p{7cm}}
\toprule
Option & Default & Introduced by & Description \\
\midrule
Mode & None & \citet{machado2018revisiting} & Different game modes, where supported \\
Difficulty & None & \citet{machado2018revisiting} & Game difficulty, where supported \\
Obs type & $rgb$ & \citet{bellemare2013ale} & Observation type from  $\lbrace rgb, grayscale, ram\rbrace$ \\
FrameSkip & 4 & \citet{bellemare2012investigating} (originally used 5) & Number of frames to skip between actions \\
Repeat action probability & 0.25 & \citet{machado2018revisiting} & Probability of actions ``sticking'' and repeating \\
Full action space & False & \citet{bellemare2013ale} & Use the full 18 actions or a minimal per-game action set \\
Continuous actions & False & \citet{farebrother2024cale} & Use continuous actions \\
Continuous action threshold & 0.5 & \citet{farebrother2024cale} & Threshold to trigger discrete events when using continuous actions \\
Max frames per episode & $108,000$ &  \citet{bellemare2013ale} & Maximum number of frames per episode \\
Sound enabled & False & \citet{towers2024gymnasium} & Add sound to the returned observations \\
\bottomrule
\end{tabular}
%\end{sc}
\end{small}
\end{center}
\vskip -0.1in
\end{table}

\begin{table}[!h]
\caption{Wrapper options controlled by three popular libraries and their defaults; libraries considered are Gymnasium \citep{towers2024gymnasium}, CleanRL \citep{huang2022cleanrl} (which uses the wrappers from StableBaselines3 \citep{stable-baselines3}), and Dopamine \citep{castro2018dopamine}.}
\label{tbl:wrapperOptions}
\vskip 0.15in
\begin{center}
\begin{small}
%\begin{sc}
\begin{tabular}{p{3.5cm}p{1.5cm}p{1.5cm}p{1.5cm}p{7cm}}
\toprule
Option & Gymnasium & SB3 / CleanRL & Dopamine & Description \\
\midrule
Max no-op & 30 & 30 & 30 & Maximum number of no-op actions taken at each reset \\
Screen size & 84x84 & 84x84 & 84x84 & Dimensions ALE observations are scaled to \\
Terminal on life loss & False & True & False & A life loss triggers a terminal signal \\
Grayscale & True & True & True & RGB observations are converted to grayscale \\
Grayscale newaxis & False & - & - & Add an extra axis to make observations 3-dimensional \\
Normalize observation & False & - & - & Normalize observations in $[0, 1)$ \\
Reward clipping & True & True & True & Clip rewards to $[-1, 1]$ \\
Fire on reset & - & True & - & Take a fire action on reset in environments that are fixed until firing \\
Max pool & True & True & True & When FrameSkip $\geq 2$, max pool the last two frames \\
\bottomrule
\end{tabular}
%\end{sc}
\end{small}
\end{center}
\vskip -0.1in
\end{table}

\section{Examples of deployed RL}
\label{app:realWorldRL}

\begin{enumerate}
    \item RLCore (https://rlcore.ai/)
    \item electric sheep (https://sheeprobotics.ai/technology/)
    \item Lyft driver/rider matching \citep{azagirre2023bettermatchdriversriders} (``implementation also requires extensive engineering work, practical RL expertise, realistic simulators, and parameter tuning.'')
    \item Ubisoft Roller Champions \citep{iskander2020reinforcementlearningagentsubisofts}
    \item Sony GT Sopy \citep{wurman2022outracing}
    \item JP Morgan hedging \citep{murray2022hedging}
    \item Siemens optimize gas turbine operation (https://www.siemens-energy.com/global/en/home/products-services/service/gt-autotuner.html)
    \item Google chip design \citep{mirhoseini2021graph}
    \item Amazon inventory management \citep{madeka2022deepinventorymanagement}
    \item Autonomous control of stratospheric balloons (no longer active) \citep{bellemare2020autonomous}
\end{enumerate}

For other examples, see the list compiled by \citet{szepesvari2023applications}.

%%%%%%%%%%%%%%%%%%%%%%%%%%%%%%%%%%%%%%%%%%%%%%%%%%%%%%%%%%%%

\end{document}